\documentclass{bmvc2k}

\usepackage{color, xcolor, comment, colortbl, caption, subcaption}
\usepackage{amsmath, amsfonts, amsthm}

\newtheorem{definition}{Definition}

\title{Zero-Shot Action Recognition from\\
Diverse Object-Scene Compositions}

\addauthor{Carlo Bretti}{bretti.carlo@gmail.com}{1}
\addauthor{Pascal Mettes}{P.S.M.Mettes@uva.nl}{1}

\addinstitution{
 University of Amsterdam\\
 Amsterdam, the Netherlands
}

\runninghead{Bretti, Mettes}{Diverse Object-Scene Compositions}

\def\eg{\emph{e.g}\bmvaOneDot}
\def\ie{\emph{i.e}\bmvaOneDot}

\def\etal{\emph{et al}\bmvaOneDot}

\definecolor{Gray}{gray}{0.9}

\DeclareMathOperator*{\argmax}{arg\,max}

\begin{document}

\maketitle

\begin{abstract}
\noindent
This paper investigates the problem of zero-shot action recognition, in the setting where no training videos with seen actions are available. For this challenging scenario, the current leading approach is to transfer knowledge from the image domain by recognizing objects in videos using pre-trained networks, followed by a semantic matching between objects and actions. Where objects provide a local view on the content in videos, in this work we also seek to include a global view of the scene in which actions occur. We find that scenes on their own are also capable of recognizing unseen actions, albeit more marginally than objects, and a direct combination of object-based and scene-based scores degrades the action recognition performance. To get the best out of objects and scenes, we propose to construct them as a Cartesian product of all possible compositions. We outline how to determine the likelihood of object-scene compositions in videos, as well as a semantic matching from object-scene compositions to actions that enforces diversity among the most relevant compositions for each action. While simple, our composition-based approach outperforms object-based approaches and even state-of-the-art zero-shot approaches that rely on large-scale video datasets with hundreds of seen actions for training and knowledge transfer.
\end{abstract}

\section{Introduction}
This work seeks to recognize actions in videos without the need for any video examples to train on. Akin to zero-shot learning in the image domain~\cite{Lampert2014Attribute-BasedCategorization,xian2018zero}, a wide range of works have shown the ability to recognize unseen actions in videos by learning a shared embedding based on seen training actions. Examples of shared action embeddings include attributes~\cite{Liu2011,Gan2016} and word embeddings~\cite{Brattoli2020,Xu2015}. Such a transfer from seen to unseen actions requires many seen actions and videos to be effective~\cite{Brattoli2020,Zhu2018TowardsRecognition} and can be biased towards specific unseen actions~\cite{Roitberg2019}. Here, we forego the need for seen actions and directly infer unseen actions in videos from other domains.

Several works have shown that unseen actions and events can be inferred by transferring visual knowledge from images and semantic knowledge from natural language. Most notably, objects have been shown to provide strong cues about which actions~\cite{Jain2015, Mettes2017, Mettes2021, Wu2016}, and events~\cite{chang2015semantic, habibian2014composite, Mettes2020ShuffledSearch} occur in videos. In such approaches for unseen actions, objects are recognized in videos and subsequently linked semantically to actions, implicitly enabling action inference~\cite{Jain2015}. Where objects provide a local video view, the larger context about the scene in which actions and objects occur is ignored. In order to take both perspectives into account, we propose a new way to recognize actions directly from objects and scenes simultaneously by modeling them as compositions.

\begin{figure}[t]
\centering
\includegraphics[width=\textwidth]{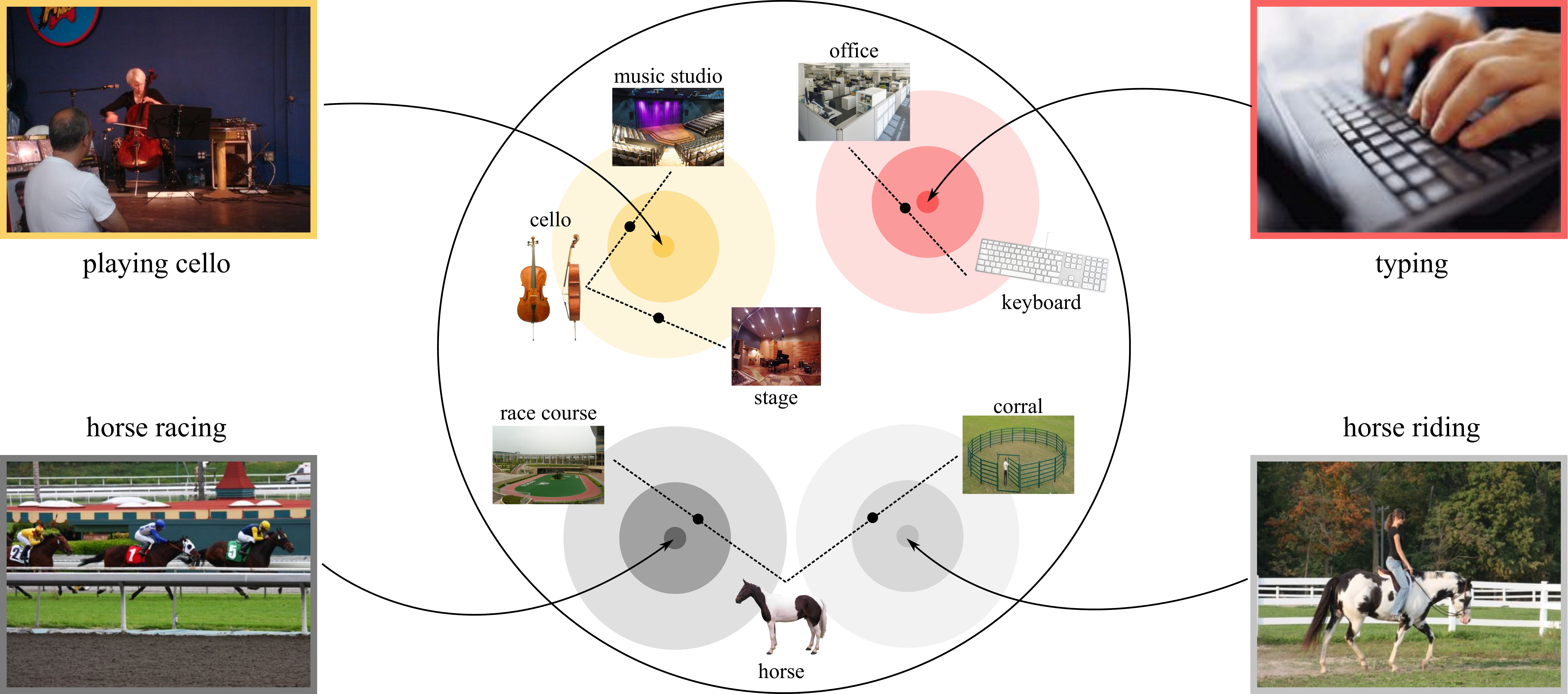}
\vspace{0.25cm}
\caption{\textbf{Intuition behind object-scene compositions for zero-shot action recognition.} Objects and scenes provide complementary views and modelling them as compositions can enrich actions like in \emph{playing cello} or \emph{typing} actions, or even distinguish actions that are identical when only considering objects, such as the case for \emph{horse racing} and \emph{horse riding}.}
\label{fig:intro}
\end{figure}

The main contribution of this work is the introduction of object-scene compositions for zero-shot action recognition. We interpret objects and scenes not as independent entities, but as a Cartesian product of all possible combinations. Figure~\ref{fig:intro} illustrates the idea behind object-scene compositions, where compositions get the best out of both knowledge sources to infer unseen actions. We furthermore describe a selection of most relevant object-scene compositions that takes the diversity amongst the objects and scenes into account. Experimental evaluation on UCF-101 and Kinetics shows the effectiveness of our composition-based approach, outperforming both object-based approaches for zero-shot action recognition and approaches that rely on hundreds of seen actions during training, despite the simplicity of the approach.
\section{Related work}
\textbf{Supervised action recognition.}
Human action recognition is a long-standing challenge in computer vision. Foundational approaches focused on the design of invariant features based on, for example, spatio-temporal interest points~\cite{Schuldt2004RecognizingApproach}, cuboids~\cite{JingenLiu2008RecognizingFeatures}, and dense trajectories~\cite{Wang2013ActionTrajectories}. Action recognition has in recent years been accelerated with advances in deep learning for videos, such as two-stream networks~\cite{Simonyan2014}, 3D convolutional networks~\cite{carreira2017quo,Ji2013}, and transformer-based networks~\cite{Girdhar2019VideoNetwork}. Deep video networks have been fueled by new large-scale videos datasets with action annotations~\cite{Kay2017, Smaira2020ADataset, Monfort2020MomentsUnderstanding, Zhao2019HACS:Localization}. While these advances make it possible to recognize up to hundreds of actions, each action requires many examples to train on. Moreover, supervised approaches do not generalize to new and unseen actions. Here, we investigate how unseen actions can still be recognized from objects and scenes.

Several supervised action approaches have previously shown the potential of knowledge about objects and scenes. Objects have shown to be effective intermediate representations for supervised action recognition~\cite{Jain2015WhatActions,li2014object} and video event detection~\cite{liu2013video,mazloom2014conceptlets,merler2011semantic}. Similarly, scenes and context information have also aided actions and events~\cite{Heilbron2015,Lien2007}. Other works investigating a combination of objects and scenes have proved their effectiveness~\cite{Wang2019CombiningRecognition, Liu2013,Ikizler-Cinbis2010}. We also seek to use objects and scenes, however we do so in the unseen setting, and we show that understanding videos as compositions of objects and scenes is fruitful.
\\\\
\textbf{Zero-shot action recognition.}
Early approaches to zero-shot action recognition followed initial success in image-based zero-shot recognition~\cite{Lampert2014Attribute-BasedCategorization} by transferring knowledge from seen to unseen actions using attributes~\cite{Liu2011,Gan2016,Zhang2015}. In attribute-based approaches, videos are mapped to an attribute space, which contains attributes such as \emph{being outdoors} and \emph{moving arms up and down}. This mapping is learned on training videos with seen actions and used during testing as an intermediate representation to generalize to unseen actions. Follow-up work advocated a direct mapping from seen to unseen actions using a shared semantic space based on word embeddings~\cite{Alexiou2016,li2016recognizing,tian2018zero,Xu2015}, using knowledge about test distributions~\cite{xu2017transductive}, feature synthesis~\cite{mishra2020zero}, or through hierarchical knowledge~\cite{long2020searching}. The current state-of-the-art seeks to construct a universal representation of actions~\cite{Zhu2018TowardsRecognition} by training a network to predict word embeddings on hundreds of seen actions~\cite{Brattoli2020}. Common among these approaches is the need for training videos with seen actions to enable a knowledge transfer to unseen actions. In this work, we seek to infer actions without the need for any seen action to train on.

In the setting without any videos to train on, Jain \etal~\cite{Jain2015} introduced an approach to classify actions based on object recognition and a semantic transfer from objects to actions. The underlying assumption of such an approach is that visually related categories are also semantically related. A similar setup has been proposed in zero-shot event detection in videos, see \eg~\cite{chang2015semantic,chang2016dynamic,habibian2014composite,Mettes2020ShuffledSearch}. An object-based setup has also been proposed for action localization by using localized object detections and prior knowledge about spatial relations between persons and objects~\cite{Mettes2017}. Most recently, object-based recognition of unseen actions has been extended by incorporating priors about multiple languages, object discrimination, and a bias towards specific object names~\cite{Mettes2021}. In this work, we follow the strict setting without any videos to train on. We build upon previous work by considering scenes in addition to objects and model these as compositions to perform action recognition. Wu \etal~\cite{Wu2016} have previously used objects and scenes for zero-shot action recognition. Their approach requires training a network to fuse object, scene, and generic features from a video to predict a seen action. From a seen action prediction, the most closely related unseen action is used as the final inference. Our composition-based approach requires no training on video examples and can directly predict any unseen action. A few works have recently investigated zero-shot compositional learning in images~\cite{mancini2021open,naeem2021learning,purushwalkam2019task}. Where these works seek to recognize unseen object-state compositions, we propose to use object-scene compositions to recognize unseen actions in videos.

\section{Unseen actions from object-scene compositions}
For our problem, we are given a set of unlabelled videos $V$ and a set of action labels $A$. We seek to find a transformation $f(v |O, S) \in A$ that assigns an action label to unseen video $v$ using two sources of prior knowledge, namely objects $O$ and scenes $S$. To that end, we first define objects and scenes as dense compositions. We then outline how to obtain zero-shot action inference from object-scene compositions and how to diversify object-scene compositions for improved performance. Figure~\ref{fig:intro} highlights the main idea, where objects and scenes are paired and used to infer which actions occur in videos through a semantic transfer from object-scene compositions to actions.

\subsection{Composition construction}
We main idea of the paper is that when using both objects and scenes to infer unseen actions, we need to consider them as compositions. We start by formally defining the compositions:
\begin{definition}
\emph{(Object-scene compositions).} Given two sets of recognition functions with corresponding semantic labels, $O$ for objects and $S$ for scenes, the object-scene compositions are defined as the Carthesian product over both sets $C = O \times S$. Each composition $c \in C$ is accompanied with a semantic representation $\phi(c)$ and a likelihood estimate $p(c | \cdot)$.
\end{definition}
\noindent
By defining objects and scenes as a Cartesian product of its pairs, we arrive at a large set of entities for which to derive semantic representations and likelihood estimates. To keep the inference tractable, we define both as linear combinations of the corresponding object and scene origin.
For a composition $c = (c_o,c_s) \in C$ consisting of one object and one scene, the semantic embedding vector is given as the summed representation over the object and scene,
\begin{equation}
\phi(c) = \phi(c_o) + \phi(c_s),
\label{embcompositions}    
\end{equation}
where $\phi(\cdot) \in \mathbb{R}^{300}$ denotes the semantic embedding function, \eg a word2vec representation~\cite{mikolov2013distributed} or sentence transformers~\cite{Wang2020MiniLM:Transformers}. To be able to perform unseen action recognition, we additionally need to be able to estimate the likelihood of each composition in a video and a semantic similarity from composition to action. For the likelihood estimation, let $\mathcal{F}_O$ and $\mathcal{F}_S$ denote deep networks pre-trained on objects $O$ and scenes $S$. Then the likelihood for video $v$ is given as:
\begin{equation}
p(c|v) = p(c_o | v ; \mathcal{F}_O) \cdot p(c_s | v ; \mathcal{F}_S).
\end{equation}
The semantic similarity between an object-scene composition $c$ and an action $a$ is given by a direct cosine similarity from their embedding~\cite{mikolov2013distributed}:
\begin{equation}
s(c, a) = \cos(\phi(c), \phi(a)).
\label{eq:casim}
\end{equation}

\subsection{Unseen action inference}
To assign an action label to an unseen video $v$, we follow object-based action literature~\cite{Jain2015,Mettes2017} and first determine the top $k$ most relevant compositions for each action based on Equation~\ref{eq:casim}. Let $C_a^k$ denote the set with the $k$ most semantically similar compositions for action $a$. Then the score for action $a$ in video $v$ is given as:
\begin{equation}
\ell(a,v) = \sum_{c' \in C_a^k} s(c', a) \cdot p(c'|v).
\label{eq:scoring}
\end{equation}
Finally, the action prediction is given as:
\begin{equation}
f(v) = \argmax_{a \in A} \ell(a,v).
\end{equation}
Similar to object-based approaches, the action inference is a relatively straightforward combination of likelihood estimation and semantic transfer. The key to our approach is that we use both objects and scenes and we consider them as compositions, which provides empirically stronger results than performing likelihood estimation and semantic transfer for objects and scenes separately.

\subsection{Diverse selection of top compositions}
For unseen action recognition, a standard approach to determine the top compositions $C_a$ for action $a$ is to select the individual compositions with the highest semantic similarity. An issue for compositions specifically is that this leads to low diversity in either the objects or scenes in the top compositions. Most notably, an object or scene that is highly semantically similar to an action will likely be over-represented in the top compositions, which leads to redundancy. We therefore seek to obtain a selection of top compositions that incorporates joint object and scene diversity.


To obtain diverse top compositions, we take inspiration from information retrieval literature~\cite{chowdhury2010introduction}, specifically Maximum Marginal Relevance~\cite{Carbonell1998TheSummaries}. For action $a$, we start by selecting the most semantically relevant composition, \ie $C_a^\star = \{\argmax_{c \in C} s(c,a) \}$. We then iteratively add the composition that is most semantically similar to action $a$ and least semantically similar to the compositions in $C_a^\star$. Hyperparameter $\lambda$ controls the relative weight of semantic similarity and diversity, where $\lambda=1$ denotes the baseline setting where only semantic similarity to the action is taken into account. The update rule for Maximum Marginal Relevance in this context is given as:
\begin{equation}
C_a^\star := C_a^\star\ \cup\ \argmax_{c' \in C \setminus C_a^\star} \bigg[ \lambda \cdot s(c', a) - (1-\lambda) \cdot \max_{c'' \in C_a^\star} s(c', c'')\bigg].
\end{equation}
We perform the above iterative update rule to populate $C_a^\star$ until it contains $k$ compositions.
\section{Experimental setup}

\subsection{Datasets}

\textbf{UCF-101} is composed of 13,320 video clips with 101 actions labels such as \emph{band marching}, \emph{ice dancing}, or \emph{pizza tossing}. The labels cover a wide range of actions, from sports to playing musical instruments and human-object interaction~\cite{Soomro2012}.
\\
\textbf{Kinetics-400} consists of 104,000 YouTube videos clips covering 400 action labels. The action labels range from individual actions such as \emph{drawing} to human-human actions such as \emph{shaking hands} and human-object interactions such as \emph{washing dishes}~\cite{Kay2017}.

\subsection{Implementation details}
\textbf{Object and scene likelihoods.}
For the object network $\mathcal{F}_o$, we follow~\cite{Mettes2021} and employ a GoogLeNet~\cite{Szegedy2015}, pre-trained on ImageNet~\cite{Deng2009} using the 12,988 object labels as per~\cite{Mettes2020ShuffledSearch}. For the scene network $\mathcal{F}_s$, we use a ResNet50~\cite{He2016DeepRecognition}, pre-trained on the Places365 dataset~\cite{Zhou2018}. For each video, we sample two frames per second and feed them to both networks. For both objects and scenes, we average their respective softmax output over all frames in the video.
\\
\textbf{Semantic Embeddings.}
We investigate two semantic embeddings for zero-shot action recognition. The first is a FastText embedding~\cite{Grave2018LearningLanguages} to allow for a direct comparison to the recent object-based approach of Mettes~\etal~\cite{Mettes2021}. We additionally investigate the potential of sentence embeddings based on the SentenceTransformers framework~\cite{Reimers2019Sentence-BERT:BERT-Networks}. Specifically, we use the pre-trained MiniLM~\cite{Wang2020MiniLM:Transformers} to generate the label embeddings. 
\\
\textbf{Code.}
The code is available \href{https://github.com/carlobretti/object-scene-compositions-for-actions}{here}.

\section{Experimental results}

For the experimental evaluation, we first perform a series of ablation studies and qualitative analyses on the proposed object-scene compositions. We then perform a comparative evaluation to the current state-of-the-art for zero-shot action recognition.

\begin{figure}[t]
\centering
\begin{subfigure}{0.49\textwidth}
\centering
\includegraphics[width=\textwidth]{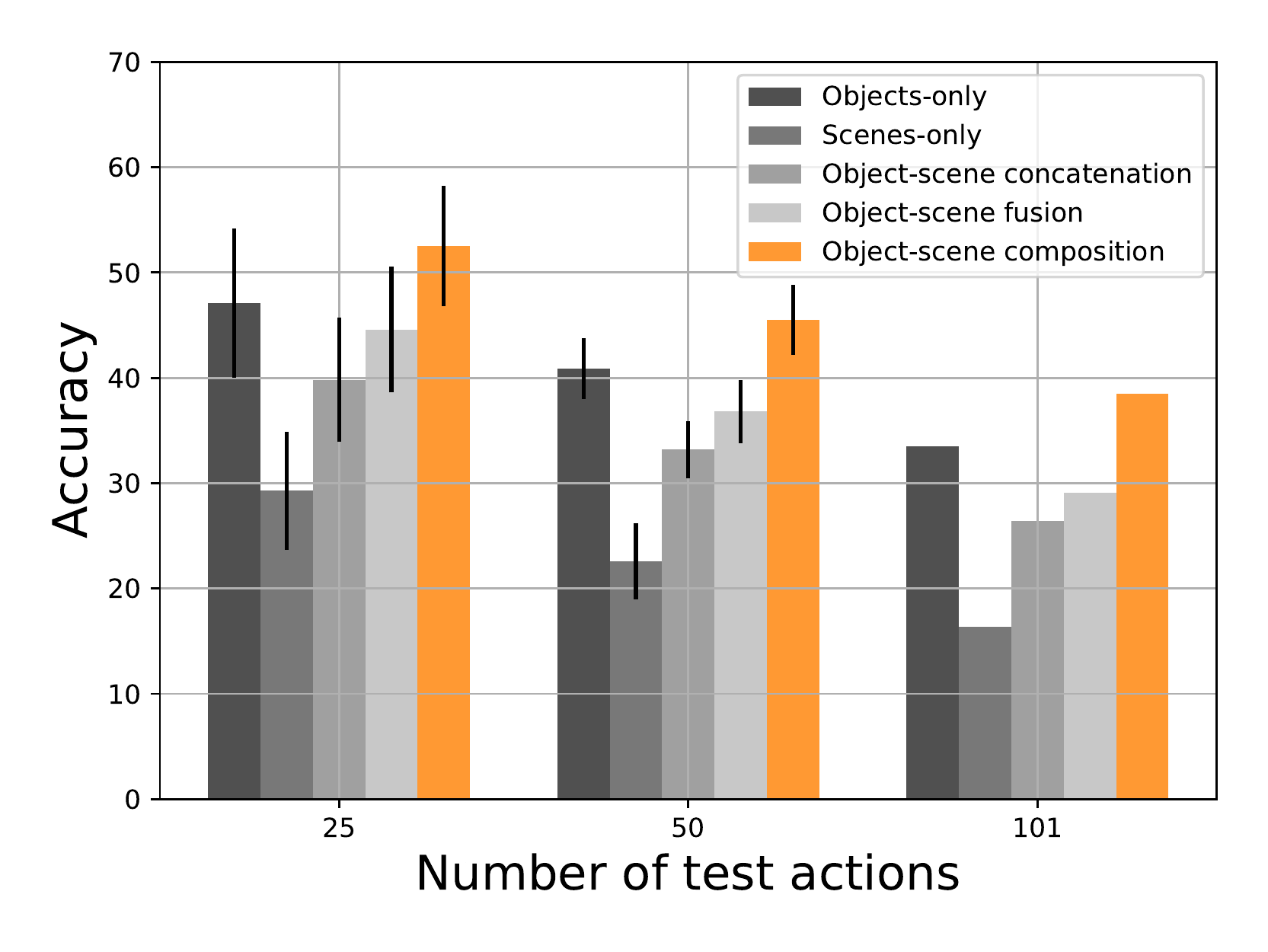}
\caption{Ablation I.}
\label{fig:ablation1}
\end{subfigure}
\begin{subfigure}{0.49\textwidth}
\centering
\includegraphics[width=\textwidth]{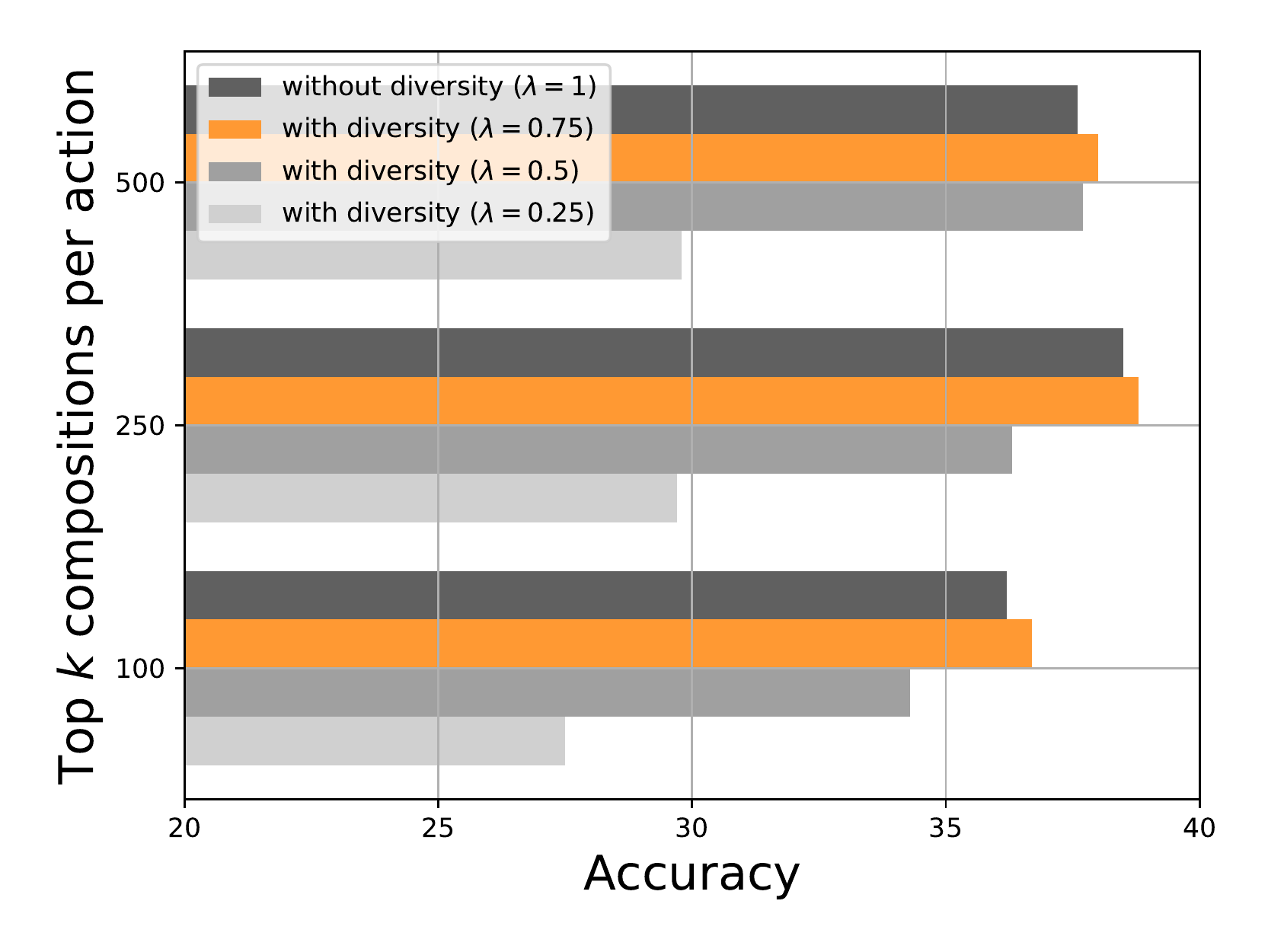}
\caption{Ablation II.}
\label{fig:ablation2}
\end{subfigure}
\vspace{0.25cm}
\caption{\textbf{Ablation studies I and II} on UCF-101. In (a), we evaluate the importance of modeling objects and scenes as compositions for zero-shot action recognition. Individually, objects are better able to differentiate actions than scenes, and standard combinations results in lower performance. Modeling objects and scenes as compositions on the other hand improves the performance, highlighting their potential. In (b) we show the effect of diverse composition selection when using all 101 actions for testing. Across all thresholds for top compositions per action, adding diversity provides a modest but consistent improvement.}
\end{figure}

\subsection{Ablation studies on object-scene compositions}
We perform the following ablation studies, to investigate: (I) the importance of compositionality when combining objects and scenes, (II) the effect of diversity in composition selection per action, (III) the effect of sentence embeddings in the semantic matching between actions and object-scene compositions, and (IV) a qualitative analysis.

\textbf{Ablation I: Importance of object-scene compositionality.}
As a starting point, we investigate the composition-based approach for combining objects and scenes. We draw a comparison to four baselines, namely one where only objects are considered, one where only scenes are considered, one where objects and scene probabilities are concatenated in a single feature, and one using a late fusion of objects and scenes by averaging their respective action scores. For the baselines, we use the top 100 objects per action as advocated in~\cite{Jain2015,Mettes2017} and the top 5 scenes per action and top 100 objects/scenes for the concatenation baseline, which work well empirically. For our approach, we use the top 250 object-scene compositions without diversity. All use FastText embeddings for semantic matching.

\begin{figure}[t]
\centering
\begin{subfigure}{\textwidth}
\centering
\includegraphics[width=0.9\textwidth,trim= 0cm 0.34cm 0cm 0cm,clip]{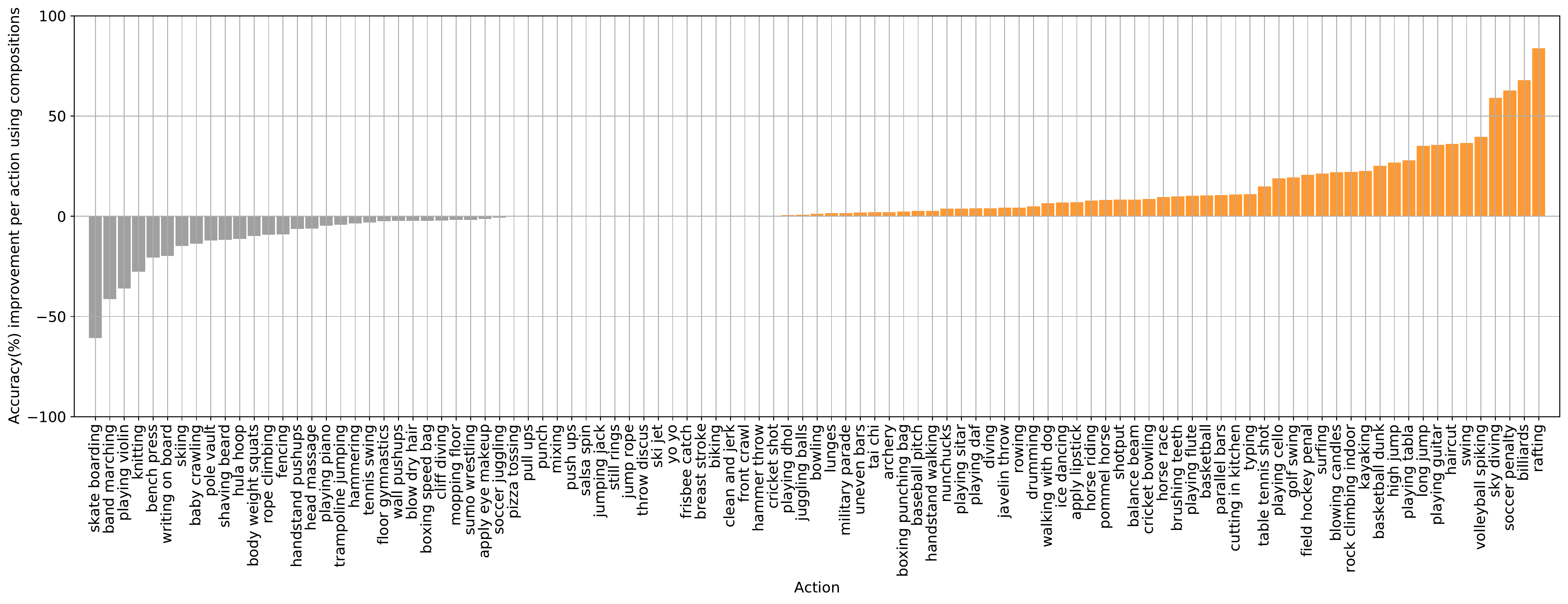}
\caption{Improvements compared to the object-only baseline.}
\label{fig:improvementOBJ}
\end{subfigure}
\begin{subfigure}{\textwidth}
\centering
\includegraphics[width=0.9\textwidth,trim= 0cm 0.34cm 0cm 0cm,clip]{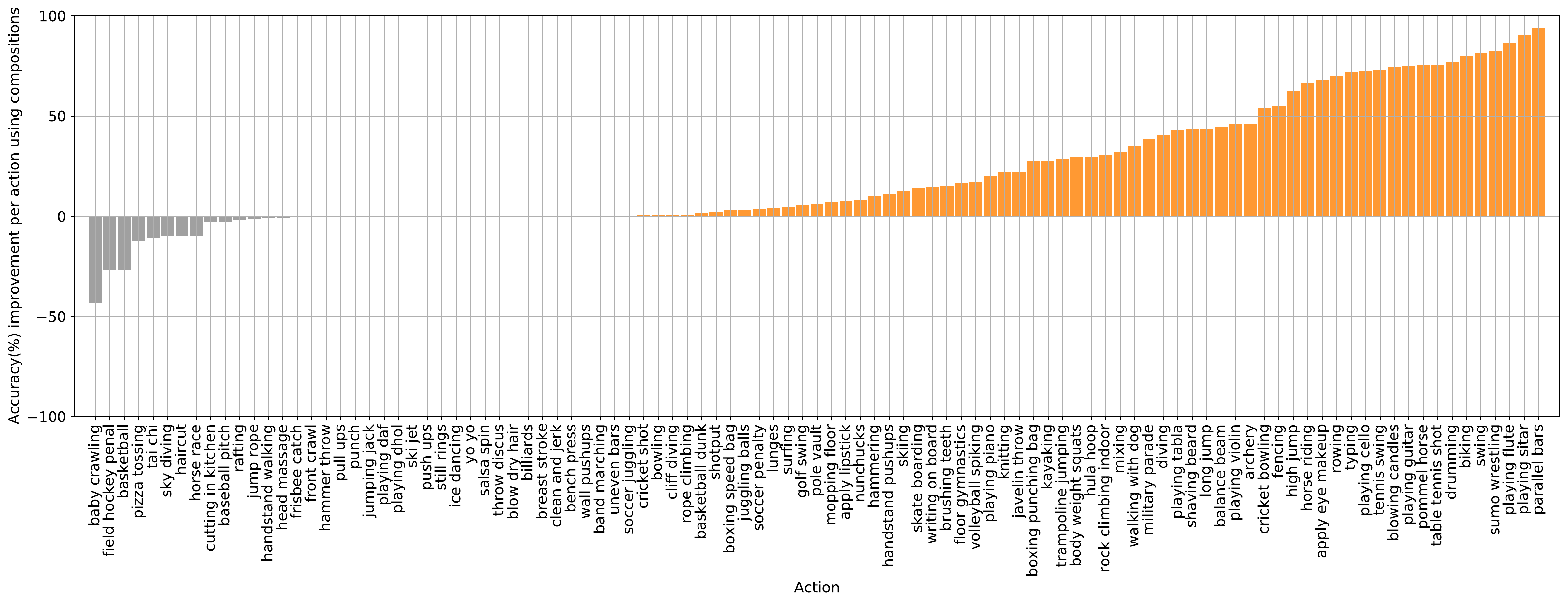}
\caption{Improvements compared to the scene-only baseline.}
\label{fig:improvementSCE}
\end{subfigure}
\vspace{0.25cm}
\caption{\textbf{Per-action improvements for our composition-based approach on UCF-101.} In (a) we show improvements compared to the object-based standard. The overall improvements are a result of large gains for scene-oriented actions (\eg \emph{soccer penalty}), while scene-independent actions do not benefit from object-scene compositions (\eg \emph{skateboarding}). In (b) we show improvements compared to the scene-only baseline. The overall improvements are a result of large gains for object-oriented actions (\eg \emph{playing sitar} or \emph{playing flute}).}
\label{fig:improvement}
\end{figure}

The results for this ablation study are shown in Figure~\ref{fig:ablation1} on UCF-101 for 25, 50, and 101 test actions. For 25 and 50 actions, we perform 10 trials where we randomly select actions for each trial and we report the mean and standard deviation over all trials. Independently, objects perform better than scenes in all three settings. The results for scenes are however far from random, indicating that there is a lot of signal in scenes for actions. Both the concatenation and late fusion of objects and scenes fail to get the best of objects and scenes. The scene-based scoring in fact has a negative effect on the object-based approach. When considering objects and scenes as compositions, however, we observe improvements across all three settings. With 101 actions for testing, the performance increases from 33.5\% for the object-based baseline to 38.5\% for our composition-based approach. We conclude that viewing objects and scenes as compositions is preferred for zero-shot action recognition.

To better understand the importance of objects and scenes as compositions for zero-shot action recognition, we have performed two additional analyses.
First, we considered objects with other objects as compositions. We made two random subsets with respectively 12,623 and 365 objects from ImageNet. On UCF-101 with 101 test actions, this approach obtains an accuracy of 32.2\% compared to 33.5\% for the object-only baseline, highlighting the importance of making compositions from diverse sources, here objects and scenes.
Second, we consider an extension of our model in which a weight for each composition is included as an additional guide for selection or scoring, similar in spirit to an object prior~\cite{Mettes2021}. The weight is defined as the cosine similarity between the object and scene embeddings. When including the weights before computing the top selection of compositions we obtain an accuracy of 30.3\%. When the weights are instead included during the scoring (\ie in Eq.\ref{eq:scoring}), we obtain an accuracy of 37.7\% compared to 38.5\% for the composition-based approach. We conclude that add a composition-based prior in the form of a weight is ineffective. Compositions themselves already contain the desired information from their textual sources.

In Figure~\ref{fig:improvementOBJ}, we show per-action improvement for all the 101 actions in UCF-101 compared to the object-only baseline. The largest improvement is obtained on \emph{rafting}, where we see an 83.7\% improvement from 4.5\% accuracy using the object-only baseline to 88.2\% using compositions. This is not surprising, since rafting occurs in a specific scene only, namely rivers with currents.
On the other hand, using compositions degrades performance on \emph{knitting} from 47\% for the object-only baseline to 22\%. This is likely due to the scene-independent nature of the action considered, as the action can take place in many scenes. Hence the stronger the link to both scenes and objects, the better the action is recognized.

\begin{table}[t]
\centering
\resizebox{0.8\linewidth}{!}{%
\begin{tabular}{lcccc}
 \hline
    & \multicolumn{4}{c}{\textbf{UCF-101}} \\
    & \multicolumn{2}{c}{50 classes} & \multicolumn{2}{c}{101 classes} \\

    & FastText & MiniLM  & FastText & MiniLM \\
 \hline
 Objects-only & 40.9 $\pm$ 2.9 &	43.6 $\pm$ 3.5 & 33.5 & 36.5\\
 Scenes-only & 22.6 $\pm$ 3.6 &  20.9 $\pm$ 2.8 &16.4 & 15.6\\
 Object-scene concatenation & 33.2 $\pm$ 2.7 & 37.8 $\pm$ 2.2 & 26.4 & 30.4\\
 Object-scene fusion & 36.8 $\pm$ 3.0 &  38.2 $\pm$ 3.0 & 29.1 & 31.1\\
 \rowcolor{Gray}
 Object-scene compositions & \textbf{45.4} $\pm$ 3.6& \textbf{45.2} $\pm$ 4.6 & \textbf{38.8} & \textbf{39.3}\\
 \hline
\end{tabular}
}
\vspace{0.25cm}
\caption{\textbf{Ablation III: Effect of sentence embeddings} on UCF-101. For all but the scene-based baseline, sentence embeddings have a positive effect on the overall performance.}
\label{tab:ablation3}
\end{table}

Figure~\ref{fig:improvementSCE} shows the improvement brought upon by using compositions rather than scenes alone. Two of the largest improvements registered are with regard to \emph{playing sitar} and \emph{playing flute}, with a 90.4\% and 86.5\% improvement respectively. This is likely due to the fact that for such actions, objects are particularly relevant and convey important information that scenes alone cannot capture. Although large improvements across most of the actions can be obtained by using compositions, for actions such as \emph{baby crawling}, \emph{field hockey penalty}, and \emph{basketball} it appears that incorporating knowledge from objects degrades performance. This is likely to be explained by a noisier prediction of object likelihoods.

\textbf{Ablation II: Effect of diverse composition selection.}
In the second ablation, we investigate the potential effect of diversifying the selection of most relevant object-scene compositions per action. We perform this ablation on UCF-101 using all 101 actions for testing with FastText for the semantic embeddings. For the diverse composition selection, we test different values of $\lambda$ as shown in Figure~\ref{fig:ablation2}.
We find that the best performance is obtained by including some notion of diversity (\ie with $\lambda = 0.75$) compared to including no diversity at all ($\lambda=1$) or to preferring diversity over relevance ($\lambda = 0.25$).
For different selections of top $k$ compositions per action, diversity provides small but consistent improvements. Throughout the rest of the paper, we will therefore incorporate diverse composition selection.

\textbf{Ablation III: Effect of sentence embeddings for semantic matching.}
Sentence embeddings have recently been shown to be beneficial for zero-shot recognition in the image domain~\cite{le2020using}. Here, we investigate their potential in the video domain. We perform a comparison between the word embeddings from FastText~\cite{Grave2018LearningLanguages} and the sentence embeddings from MiniLM~\cite{Wang2020MiniLM:Transformers} on UCF-101 using 50 and 101 test actions. The results are shown in Table~\ref{tab:ablation3}. For all approaches except the scene-only baseline, using sentence embeddings provides a direct improvement. On 101 actions, the object-only baseline improves from 33.5\% to 36.5\%, while our composition-based approach improves from 38.8\% to 39.3\%. We conclude that sentence embeddings are also beneficial for matching objects and compositions to actions.
\begin{figure}[t]
\centering
\begin{subfigure}{0.30\textwidth}
\centering
\includegraphics[width=\textwidth, page=1, trim= 25.5cm 14cm 25.5cm 8cm,clip]{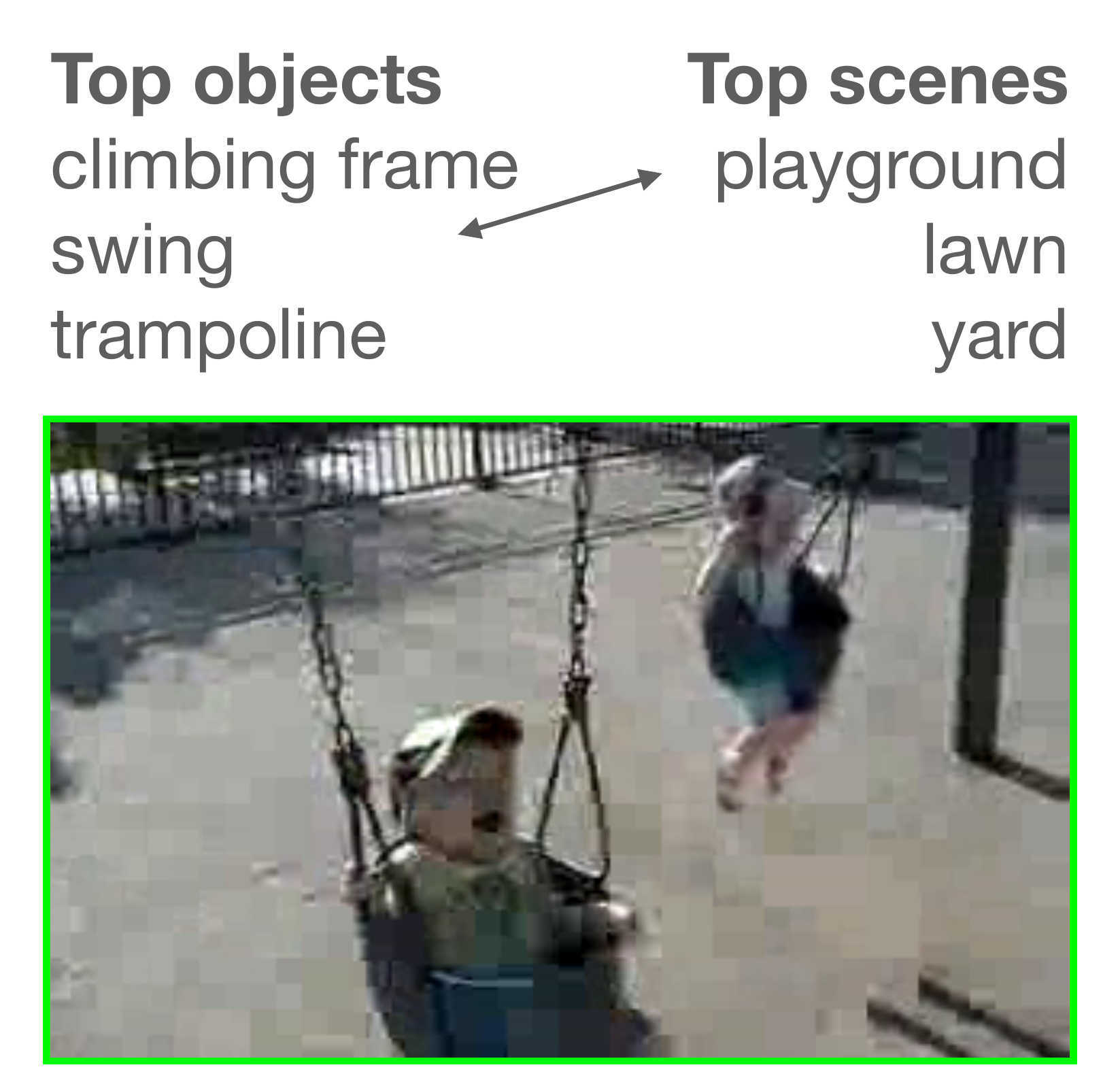}
\caption{Swinging.}
\label{fig:qualswing}
\end{subfigure}
\begin{subfigure}{0.30\textwidth}
\centering
\includegraphics[width=\textwidth, page=2, trim= 25.5cm 14cm 25.5cm 8cm,clip]{images/qualitative.pdf}
\caption{Sky Diving.}
\label{fig:qualskydiving}
\end{subfigure}
\begin{subfigure}{0.30\textwidth}
\centering
\includegraphics[width=\textwidth, page=3, trim= 25.5cm 14cm 25.5cm 8cm,clip]{images/qualitative.pdf}
\caption{Skateboarding.}
\label{fig:qualskate}
\end{subfigure}
\vspace{0.25cm}
\caption{\textbf{Qualitative results for object-scene compositions} on UCF-101. \emph{Swinging} and \emph{sky diving} have ambiguous relations to objects and scenes by themselves, while their top composition (arrows) provides a strong and disambiguated relation to their action. A composition-based approach is not beneficial when the object or scene are independent of the action, as is the case for the \emph{skateboarding} example.}
\label{fig:qual}
\end{figure}

\textbf{Ablation IV: Qualitative analysis.} Figure \ref{fig:qual} shows three example videos from UCF-101. For Figure \ref{fig:qualswing}, we can see that compositions are an effective tool, as \emph{swinging} can effectively be understood as a composition of the object \emph{swing} used in a certain context,~\ie a \emph{playground}. Indeed, \emph{swinging} directly benefits from compositions, as per Figure~\ref{fig:improvement}.
In Figure \ref{fig:qualskydiving}, \emph{dive} and \emph{sky} are deemed a top object-scene composition for the action \emph{sky diving}. For this action, objects and scenes in isolation are ambiguous, while their composition removes this ambiguity.
In Figure \ref{fig:qualskate}, we also show a failure case for \emph{skateboarding}, which is object-dependent but scene-independent.


\begin{table}[t]
\centering
\resizebox{0.625\linewidth}{!}{%
\begin{tabular}{lccc}
    \hline
    & \multicolumn{3}{c}{\textbf{UCF-101}} \\
    & \multicolumn{2}{c}{nr. actions} & accuracy\\
    & train & test & \\
    \hline
    Zhu~\etal~\cite{Zhu2018TowardsRecognition}\ \ {\footnotesize CVPR'18} & 200 & 101 & 34.2\\
    Brattoli~\etal~\cite{Brattoli2020}\ \ {\footnotesize CVPR'20} & 664 & 101 & 37.6\\
    \hline
    Jain~\etal~\cite{Jain2015}\ \ {\footnotesize ICCV'15} & - & 101 & 30.3\\
    Mettes and Snoek~\cite{Mettes2017}\ \ {\footnotesize ICCV'17} & - & 101 & 32.8\\
    Mettes~\etal~\cite{Mettes2021}\ \ {\footnotesize IJCV'21} & - & 101 & 36.3\\
    \rowcolor{Gray}
    \textbf{This paper} w/ word embeddings & - & 101 & \textbf{38.8}\\
    \rowcolor{Gray}
    \textbf{This paper} w/ sentence embeddings & - & 101 & \textbf{39.3}\\
    \hline
\end{tabular}
}
\vspace{0.25cm}
\caption{\textbf{Comparative evaluation on UCF-101.} Both compared to the object-based state-of-the-art~\cite{Mettes2021} and compared to the overall best performing approach that uses 664 seen Kinetics actions for training~\cite{Brattoli2020}, we obtain higher accuracies. We obtain state-of-the-art results with word embeddings and obtain further improvements with sentence embeddings.}
\label{tab:ucf101sota}
\end{table}

\begin{table}[t]
\centering
\resizebox{0.575\linewidth}{!}{%
\begin{tabular}{lccc}
 \hline
    & \multicolumn{3}{c}{\textbf{Kinetics}} \\
    & \multicolumn{3}{c}{Number of test actions} \\
  & 25 & 100 & 400\\
 \hline
 Mettes~\etal~\cite{Mettes2017} as in~\cite{Mettes2021}& 21.8 $\pm$ 3.5	& 10.8 $\pm$ 1.0 &	6.0\\
 Mettes~\etal~\cite{Mettes2021} & 22.0 $\pm$ 3.7	& 11.2 $\pm$ 1.0 &	6.4\\
 Objects-only &18.3 $\pm$ 2.5 & 10.5 $\pm$ 0.9 & 5.8\\
 Scenes-only &28.2 $\pm$ 6.0 & 14.2 $\pm$ 1.2 & 7.3\\
 \rowcolor{Gray}
 \textbf{This paper} & \textbf{29.7} $\pm$ 5.0 & \textbf{18.0} $\pm$ 1.1 & \textbf{9.4}\\
 \hline
\end{tabular}
}
\vspace{0.25cm}
\caption{\textbf{Comparative evaluation on Kinetics.} In the Kinetics dataset, scenes are more informative for actions than objects, as indicated by the higher classification performance. The composition between objects and scenes provides the best overall results.}
\label{tab:kincomp}
\end{table}

\subsection{Comparative evaluation}
We perform a comparison to the current state-of-the-art in zero-shot action recognition on UCF-101 in the most challenging setting, namely using all 101 actions for testing. We also perform a comparison on Kinetics, which has recently been performed in~\cite{Mettes2021}.

\textbf{Comparison on UCF-101.} In Table~\ref{tab:ucf101sota}, we compare our approach to the current state-of-the-art in zero-shot action recognition. The most direct comparison is to object-based action recognition approaches~\cite{Jain2015,Mettes2017,Mettes2020ShuffledSearch}, which do not use any training videos with seen actions. Where Mettes \etal~\cite{Mettes2021} obtain an accuracy of 36.3\%, we improve the results to 38.8\% with the same word embeddings and further boost the accuracy to 39.3\%  with sentence embeddings. We also include a comparison to Zhu \etal~\cite{Zhu2018TowardsRecognition} and Brattoli~\etal~\cite{Brattoli2020}, which perform zero-shot action recognition by training on hundreds of seen actions and performing a semantic transfer from seen to unseen actions. Compare to the state-of-the-art approach by Brattoli~\etal~\cite{Brattoli2020}, we improve the overall accuracy from 37.6\% to 39.3\%\, highlighting the effectiveness of object-scene compositions for zero-shot action recognition.

\textbf{Comparison on Kinetics.} We also perform a comparative evaluation on Kinetics, based on the zero-shot protocol recently outlined in~\cite{Mettes2021}. The object-only baseline follows an approach comparable to that of Jain~\etal\cite{Jain2015}. We find scenes to be slightly more informative than objects. Modeling objects and scenes as compositions remains directly beneficial. Where Mettes \etal~\cite{Mettes2021} obtain a mean accuracy of 11.2\% with 100 actions, our mean accuracy is 18.0\%. On the full 400-way classification, our approach improves the scores from 6.4\% to 9.4\%, restating the effectiveness of our composition-based approach.

\section{Conclusions}
This work advocates objects and scenes as compositions for zero-shot action recognition. Where the current standard is to transfer knowledge from objects when no training videos are available, we propose a Cartesian product of all objects and scenes as basis for inferring unseen actions. Due to the homogeneous nature of such a set, we find that actions benefit from diversity during composition selection. Experimentally, we show that object-scene compositions provide a simple yet effective approach for zero-shot action recognition, especially in settings where relying on objects or scenes is ambiguous. Comparisons on UCF-101 and Kinetics show that our composition-based approach outperforms the current object-based standard and even recent works that rely on training on hundreds of seen actions.

\bibliography{egbib}
\end{document}